\documentclass{article}
\usepackage{kr}
\usepackage{times}
\usepackage{url}
\usepackage[utf8]{inputenc}
\usepackage[small]{caption}
\usepackage{graphicx}
\usepackage{amsmath}
\usepackage{amsthm}
\usepackage{booktabs}
\usepackage{amssymb}
\usepackage{multirow}
\usepackage{mathrsfs}
\usepackage{stmaryrd}
\usepackage{listings}
\usepackage[usenames,dvipsnames,svgnames,table]{xcolor}
\usepackage[unicode,pdfmenubar,linktoc=all,hidelinks,bookmarks, pdftex]{hyperref}
\usepackage{xspace}
\usepackage{graphicx}
\usepackage{csquotes}
\usepackage{xparse}
\usepackage{booktabs}
\usepackage{enumerate}
\usepackage[kerning,spacing]{microtype}
\usepackage{comment}
\usepackage{tikzlings}

\usepackage{amssymb}
\usepackage{multirow}
\usepackage{mathrsfs}
\usepackage{stmaryrd}
\usepackage{listings}
\usepackage{subcaption}
\usepackage[usenames,dvipsnames,svgnames,table]{xcolor}
\usepackage[ruled,vlined]{algorithm2e}
\usepackage{enumerate}
\usepackage{xspace}
\usepackage{csquotes}
\usepackage{xparse}
\usepackage{wasysym}
\usepackage[kerning,spacing]{microtype}

\usepackage{tikz}
\usetikzlibrary{fadings}
\usetikzlibrary{shapes,decorations,positioning}
\usetikzlibrary{fadings,shapes,decorations,positioning,calc}
\usetikzlibrary{decorations.markings,decorations.pathreplacing}
\usetikzlibrary{shapes.geometric,automata,positioning}
\usetikzlibrary{shadows}
\usetikzlibrary{calc}
\usetikzlibrary{decorations,decorations.pathmorphing,decorations.pathreplacing}
\usetikzlibrary{calc}

\newtheorem{theorem}{Theorem}
\newtheorem{example}[theorem]{Example}
\newtheorem{definition}[theorem]{Definition}
\newtheorem{proposition}[theorem]{Proposition}

\newtheorem{corollary}[theorem]{Corollary}

\makeatletter
\newcommand{\textlabel}[2]{%
    \protected@edef\@currentlabel{#1}%
    \phantomsection%
    #1\label{#2}%
}
\makeatother

\DeclareMathOperator{\Cn}{Cn}

\newcommand{\setAllES}{\ensuremath{\mathcal{E}}}
\DeclareMathOperator{\beliefSymb}{Bel}
\newcommand{\beliefsOf}[1]{\ensuremath{\beliefSymb(#1)}}

\newcommand{\modelsOf}[1]{\ensuremath{\llbracket #1\rrbracket}}
\newcommand{\minOf}[2]{\ensuremath{\min(#1,#2)}}
\newcommand{\negOf}[1]{\ensuremath{\neg{#1}}}
\newcommand{\epistemicSpace}{\ensuremath{\mathbb{E}}}
\newcommand{\logicLanguage}{\ensuremath{\mathcal{L}}}
\newcommand{\revision}{\ensuremath{\star}}
\newcommand{\contraction}{\ensuremath{\div}}

\newcommand{\tuple}[1]{\langle#1\rangle}

\newcommand{\ksIF}{\text{if }}
\newcommand{\ksTHEN}{\text{, then }}
\newcommand{\ksOtherwise}{\text{otherwise}}

\newcommand{\logicBeliefSet}{\logicLanguage^{\beliefSymb}}

\newcommand{\myparagraph}[1]{\par\smallskip\noindent\textbf{#1.}}

\title{The Realizability of Revision and Contraction Operators in Epistemic Spaces}
\author{
Kai Sauerwald
\and
Matthias Thimm
\affiliations
Artifical Intelligence Group, FernUniversität in Hagen, 58084 Hagen, Germany
\emails
\{kai.sauerwald, matthias.thimm\}@fernuni-hagen.de
}

\begin{document}
    \frenchspacing

\maketitle
\pagestyle{plain}

\begin{abstract}
This paper studies the realizability of belief revision and belief contraction operators in epistemic spaces. We observe that AGM revision and AGM contraction operators for epistemic spaces are only realizable in precisely determined epistemic spaces. We define the class of linear change operators, a special kind of maxichoice operator. When AGM revision, respectively, AGM contraction, is realizable, linear change operators are a canonical realization.
\end{abstract}

\section{Introduction}
The area of belief change deals with the adaption of agents’ beliefs in the light of new information from a fundamental perspective. 
A general novel framework for considering belief change is the framework of belief change operators for epistemic spaces \cite{KS_SchwindKoniecznyPinoPerez2022}.
In this framework, belief change happens on a specific defined set of epistemic states. 
The belief change operators are global mathematical objects that define
how each state changes in the light of arbitrary new information.
An advantage of this framework is that it recognizes the varying nature of epistemic states of individual agents, which is given, e.g., due to the individual capabilities of the agents.
Many other frameworks do not allow such a differentiation, because they (implicitly) assume that all belief sets are available.

In this paper, we study the \emph{realizability} of belief change operators: the question of whether there exists a belief change operator that satisfies a given set of postulates under a given set of assumptions.
This problem been addressed for many types of belief change operators in various frameworks \cite{KS_FermeHansson2018,KS_FalakhRudolphSauerwald2022,KS_RibeiroWassermannFlourisAntoniou2013} and are an ongoing research subject, but has been not considered for the expressive framework of epistemic spaces.

The generality of the framework  for epistemic spaces and the global nature of belief change operators in this framework has consequences for the problem of realizability.
To check realizability, one must consider all belief changes on all epistemic states (available in the considered epistemic space) together and decide if all those belief changes do not interfere with each other.
This is different to deliberations about the realizability of belief change operators considered in many other frameworks, where a realizability is about the changes in one particular belief set or sequences of changes. %

In the following, we list the main content of this paper, which also includes the main contributions of this submission:
\begin{itemize}
    \item{[Non-Existence]} 
    Our first observation is that AGM revision and AGM contraction operators \cite{KS_AlchourronGaerdenforsMakinson1985} for epistemic states \cite{KS_DarwichePearl1997} do  not exist in all epistemic spaces.
    Furthermore, the corresponding epistemic spaces do not coincide.

    \item{[Realizability of Revision and Contraction]} A precise characterization of those epistemic spaces is given, where any AGM revision operators, respectively, any AGM contraction operators, exist at all.
    
    \item{[Realizability of Operators]} 
    We consider then realizability of concrete belief revision and contraction operator approaches: full meet, maxichoice \cite{KS_AlchourronMakinsion1982} and newly defined linear change operators.
    For contraction, we obtain the realizability of all these operators coincides with realizability of contraction.
    When considering realizability of revision operators, we observe that realizability of general revision, linear revision and maxichoice revision coincide.
For full meet revision 
 the realizibility is different. 

    \item{[Non-Interdefinability of Revision and Contraction]} 
     A consequence of our investigation is that there are epistemic space where one of AGM revision and AGM contraction is realizable and the other type of operator not. This shows that in some epistemic spaces revision and contraction are not interdefinable, which is an extension to a result obtained by \citeauthor{KS_KoniecznyPinoPerez2017} (\citeyear{KS_KoniecznyPinoPerez2017}).
\end{itemize}
We have proofs for all results given in this paper, which can be found in the accompanied
	supplementary material. 

\section{Propositional Logic, Minima and Orders}
\label{sec:background}
Let $ \Sigma=\{a,b,c,\ldots\} $ be a finite set of propositional atoms and let $ \logicLanguage $ be a propositional language over $ \Sigma $. 
The set of propositional interpretations is denoted by $ \Omega $.
Propositional entailment is denoted by $ \models $ and the set of models of $ \alpha $ with $ \modelsOf{\alpha} $.
For \( L\subseteq \logicLanguage \),  entailment and model sets are defined as usual, i.e., \( \modelsOf{L}=\bigcap_{\alpha\in L} \modelsOf{\alpha}   \) and \( L\models \beta \) if for all \( \alpha\in L \) holds \( \alpha \models \beta \).
We define \( Cn(L)= \{ \beta\mid L \models \beta \} \) and we define \( L+\alpha = \mathrm{Cn}(L\cup\{\alpha\}) \). Furthermore, \( L \) is called deductively closed if \( L = \mathrm{Cn}(L) \) and \( \logicBeliefSet \) is the set of all deductively closed sets.
Given $ \Omega'\subseteq \Omega $ and a total preorder $ {\leq} \subseteq  \Omega \times \Omega $ (total and transitive relation), we denote with 
$ {\min(\Omega',\leq)}=\{ \omega\in\Omega' \mid  \omega \leq \omega' \text{ for all } \omega'\in\Omega' \} $
the set of all minimal interpretations of $ \Omega' $ with respect to \( {\leq} \).
A linear order \( {\ll} \subseteq \Omega\times\Omega \) is a total preorder that is antisymmetric.

\section{Epistemic~Spaces, Revision and Contraction}
\label{sec:background_bc}
We consider the background on epistemic spaces, AGM revision and AGM contraction.
In this work, we model agents by the means of logic. 
Deductive closed sets of formulas,  which we denote from now as \emph{belief set}, represent deductive capabilities; agents are assumed to be
perfect reasoners.
The interpretations represent worlds that the agent is capable to imagine.
The following notion describes the space of epistemic possibilities of an agent's mind in a general way.

\begin{definition}[\citeauthor{KS_SchwindKoniecznyPinoPerez2022} \citeyear{KS_SchwindKoniecznyPinoPerez2022}; adapted]\label{def:epistemic_space}
    A tuple 
    \( \epistemicSpace
    =
    \langle
    \setAllES,\beliefSymb\rangle  \) is called an \emph{epistemic space} if $ \setAllES $ is a non-empty set 
    and $ \beliefSymb : \setAllES \to \logicBeliefSet $.
\end{definition}
We call the elements of \( \setAllES \) \emph{epistemic states} and use \( \modelsOf{\Psi} \) as shorthand for \( \modelsOf{\beliefsOf{\Psi}} \). %
Definition~\ref{def:epistemic_space}  differs from the definition given by \citeauthor{KS_SchwindKoniecznyPinoPerez2022} {(\citeyear{KS_SchwindKoniecznyPinoPerez2022})} insofar that we do \emph{not} exclude inconsistent belief sets and forbid emptiness of \( \setAllES \).
Belief change operators for an epistemic space \( \epistemicSpace \) are global objects, functions on all epistemic states in the mathematical sense.
\begin{definition}\label{def:belief_change_operator_es}
	Let \( \epistemicSpace
	=
	\langle
	\setAllES,\beliefSymb\rangle  \) be an epistemic space.
	A \emph{belief change operator for $ \epistemicSpace $}  is a function $ \circ : \setAllES \times \logicLanguage \to 
	\setAllES $.
\end{definition}
In general, operators from Definition~\ref{def:belief_change_operator_es} could behave arbitrarily, and, of course, for considering revision and contraction, we have to add additional constraints, which we will consider in the following.

\myparagraph{AGM Belief Revision for Epistemic Spaces}
Revision operators incorporating new beliefs into an agent's belief set, consistently, whenever this is possible.
We use an adaption of the AGM postulates for revision \cite{KS_AlchourronGaerdenforsMakinson1985}
for epistemic states {\cite{KS_DarwichePearl1997}, which is inspired by the approach of \citeauthor{KS_KatsunoMendelzon1992}  {(\citeyear{KS_KatsunoMendelzon1992})}.
    \pagebreak[3]
    \begin{definition}\label{def:agm_es_revision}
        Let \( \epistemicSpace = \langle \setAllES,\beliefSymb \rangle  \) be an epistemic space.
        A belief change operator \( \revision \) for \( \epistemicSpace \) is called an \emph{(AGM) revision operator  for \( \epistemicSpace \)} if the following postulates are satisfied {\cite{KS_DarwichePearl1997}}: %
        \begin{description}
            \item[\normalfont(\textlabel{R1}{pstl:R1})] 
                \( \alpha \in  \beliefsOf{\Psi \revision \alpha}  \)
            \item[\normalfont(\textlabel{R2}{pstl:R2})] 
                \(\beliefsOf{\Psi \revision \alpha} = {\beliefsOf{\Psi} + \alpha } \) if \( \beliefsOf{\Psi} + \alpha\) is consistent
            \item[\normalfont(\textlabel{R3}{pstl:R3})] 
                If \(  \alpha \text{ is consistent, then }  \beliefsOf{\Psi \revision \alpha} \text{ is consistent}  \)
            \item[\normalfont(\textlabel{R4}{pstl:R4})] 
                If \( \alpha \equiv \beta \ksTHEN  \beliefsOf{\Psi \revision \alpha} = \beliefsOf{\Psi \revision \beta}  \)
            \item[\normalfont(\textlabel{R5}{pstl:R5})] 
                \( \beliefsOf{\Psi \revision (\alpha\land\beta)} \subseteq \beliefsOf{\Psi \revision \alpha}+\beta \)
            \item[\normalfont(\textlabel{R6}{pstl:R6})] 
                If \(  \beliefsOf{\Psi\revision\alpha} + \beta \) is consistent,\\\null\hspace{3em}then \(  \beliefsOf{\Psi \revision \alpha}+\beta \subseteq \beliefsOf{\Psi \revision (\alpha\land\beta)}  \)
        \end{description}\pagebreak[3]
    \end{definition}
    AGM revision is well-known for aiming at establishing a minimal change of the prior beliefs when revising. 
    This is carried mainly by \eqref{pstl:R2} with respect to the beliefs of an agent.
    Moreover, \eqref{pstl:R5} and \eqref{pstl:R6} are postulates about rational choice that ensure relational minimal change \cite{KS_AlchourronGaerdenforsMakinson1985,KS_KatsunoMendelzon1992,KS_FermeHansson2018}; an interpretation that is also used in other areas \cite{KS_Sen1971}.
    For a detailed discussion of the postulates \eqref{pstl:R1}--\eqref{pstl:R6} we refer to Gärdenfors (\citeyear{KS_Gaerdenfors1988}).
    In the remaining parts of this paper, we sometimes write \emph{revision operator} instead of \emph{AGM revision operator}.
    
    \myparagraph{AGM Belief Contraction for Epistemic Spaces}
    Contraction is the process of withdrawing beliefs, without adding new beliefs.
    Postulates for AGM contraction \cite{KS_AlchourronGaerdenforsMakinson1985} where adapted by \citeauthor{KS_CaridroitKoniecznyMarquis2015} (\citeyear{KS_CaridroitKoniecznyMarquis2015}) to the setting of propositional logic.
    An adapted version of these postulates for AGM contraction operators for epistemic spaces from \citeauthor{KS_KoniecznyPinoPerez2017} (\citeyear{KS_KoniecznyPinoPerez2017})  is given in the following.
    \begin{definition}[Adapted, {\citeauthor{KS_KoniecznyPinoPerez2017}}, {\citeyear{KS_KoniecznyPinoPerez2017}}]
        Let \( \epistemicSpace = \langle \setAllES,\beliefSymb \rangle  \) be an epistemic space.
        A belief change operator \( \contraction \) for \( \epistemicSpace \) is called an \emph{(AGM) contraction operator for \( \epistemicSpace \)} if the following postulates are satisfied:
        \begin{description}
            \item[\normalfont(\textlabel{C1}{pstl:C1})]  \( \beliefsOf{\Psi \contraction \alpha} \subseteq \beliefsOf{\Psi}  \)
            
            \item[\normalfont(\textlabel{C2}{pstl:C2})] \( \text{If }  \alpha\notin\beliefsOf{\Psi} \ksTHEN \beliefsOf{\Psi}\subseteq \beliefsOf{\Psi \contraction\alpha} \)
            
            \item[\normalfont(\textlabel{C3}{pstl:C3})] \( \text{If } \alpha \not\equiv \top \ksTHEN  \alpha \notin \beliefsOf{\Psi  \contraction  \alpha} \)
            
            \item[\normalfont(\textlabel{C4}{pstl:C4})] \( \beliefsOf{\Psi} \subseteq \beliefsOf{\Psi  \contraction  \alpha} + \alpha \)
            
            \item[\normalfont(\textlabel{C5}{pstl:C5})] \( \text{If } \alpha \equiv\beta \ksTHEN \beliefsOf{\Psi  \contraction  \alpha} = \beliefsOf{\Psi  \contraction  \beta} \)
            
            \item[\normalfont(\textlabel{C6}{pstl:C6})] \( \beliefsOf{\Psi \contraction \alpha} \cap \beliefsOf{\Psi \contraction \beta} \subseteq \beliefsOf{\Psi  \contraction  (\alpha\land\beta)} \)
            
            \item[\normalfont(\textlabel{C7}{pstl:C7})] \( \text{If } \beta\notin\beliefsOf{\Psi   \contraction  (\alpha\land\beta)}  \),\\\null\hspace{3em}then \( \beliefsOf{\Psi  \contraction  (\alpha\land\beta)} \subseteq \beliefsOf{\Psi \contraction \beta} \)
        \end{description}
    \end{definition}
    As in the case of revision, AGM contraction is well-known for aiming at establishing a minimal change of the prior beliefs when contracting, which is carried mainly by \eqref{pstl:C1}, \eqref{pstl:C2}, \eqref{pstl:C6} and \eqref{pstl:C7}. 
    Again, we refer to Gärdenfors (\citeyear{KS_Gaerdenfors1988}) for a detailed discussion of the postulates \eqref{pstl:C1}--\eqref{pstl:C7}.
    In the remaining parts of this paper we sometimes write \emph{contraction operator} instead of \emph{AGM contraction operator}.

\section{Realizability of Contraction Operators}
\label{sec:realization_contraction}
We define realizability of AGM contraction as notion that captures whether there is  any contraction  operator.
\begin{definition}[Contraction Realizability]\label{def:realizContraction}
   We say that \emph{AGM contraction is realizable} in an epistemic space \( \epistemicSpace \), if there exists an AGM contraction operator for \( \epistemicSpace \).
\end{definition}

The following proposition fully captures those epistemic spaces \( \epistemicSpace \) for which an AGM contraction operator for \( \epistemicSpace \) exists.
\begin{theorem}\label{prop:min_es_for_agmcontraction}
	Let \( \epistemicSpace = \langle \setAllES,\beliefSymb \rangle  \) be an epistemic space.
	AGM contraction is realizable in \( \epistemicSpace \) if and only if \( \epistemicSpace \) satisfies:
	\begin{description}
		\item[\normalfont(\textlabel{ZC}{pstl:AGM-ES3})] For each \( \Psi\in\setAllES \) and for each \( M \subseteq \Omega \) with \( \modelsOf{\Psi}\subseteq M \)\\\null\hspace{2.25em} there exists \( \Psi_{M}\in\setAllES \) such that \( \modelsOf{\Psi_{M}}=M \).
	\end{description}
\end{theorem}
The postulate \eqref{pstl:AGM-ES3} describes that agents are always capable of being more undecided about the state of the world. 
Clearly, Theorem~\ref{prop:min_es_for_agmcontraction} implies that contraction operators do not exist in every epistemic space. 
For specific types of contraction operators, the realizability could be even more strict.
To that end, we now define some basic types of contraction operators for epistemic states that are inspired by classical approaches to contraction \cite{KS_Hansson1999}.
\begin{definition}\label{def:contraction_fullmeet_maxichoice}
    Let \( \epistemicSpace = \langle \setAllES,\beliefSymb \rangle  \) be an epistemic space and let \( \contraction \) be a contraction operator for \( \epistemicSpace \).
        We say \( \contraction \) is a \emph{full meet contraction operator for \( \epistemicSpace \)} if for all \( \Psi\in\setAllES \) and \( \alpha\in\logicLanguage \) holds:
        \begin{equation*} %
            \modelsOf{\Psi\contraction\alpha}=\begin{cases}
                \modelsOf{\Psi}  & \ksIF \modelsOf{\Psi}\cap\modelsOf{\negOf{\alpha}} \neq \emptyset\\
                \modelsOf{\Psi}\cup\modelsOf{\negOf{\alpha}}  & \ksOtherwise
            \end{cases}
        \end{equation*}
        We say \( \contraction \) is a \emph{maxichoice contraction operator for \( \epistemicSpace \)} if for each \( \Psi \,{\in}\, \setAllES \) exists a linear order \( \ll_{\!\Psi} \) such that for all \( \alpha \,{\in}\, \logicLanguage \) holds:  
        \begin{equation*} %
            \modelsOf{\Psi\contraction\alpha} = \begin{cases}
                \modelsOf{\Psi}  & \ksIF \modelsOf{\Psi}\cap\modelsOf{\negOf{\alpha}} \neq \emptyset\\
                \modelsOf{\Psi}\cup\minOf{\modelsOf{\negOf{\alpha}}}{\ll_\Psi} & \ksOtherwise
            \end{cases}
        \end{equation*}
\end{definition}
\begin{figure}[t]
\centering
\begin{tikzpicture}
	\tikzstyle{esnode}=[circle,draw,fill=gray!15,text=black,minimum width=2.5em,minimum height=1.75em]
	\tikzstyle{myedge}=[->,thick,-{latex}]
	
	\tikzstyle{myedgenode}=[inner sep=0.2ex,sloped]

	\node (psiAB) [esnode,anchor=west] at (3,0.8) {\(\Psi_{a}\)};
	\node (psiANB) [esnode,anchor=west] at (3,-0.8) {\(\Psi_{\overline{a}}\)};
	\node (psiT) [esnode,anchor=west] at (6,0) {\(\Psi_{\top}\)};
	
	\draw (psiAB) edge [loop left,thick] node {$*$}  (psiA) ;
	\draw (psiANB) edge [loop left,thick] node {$*$}  (psiA) ;
	\draw (psiT) edge [loop right,thick] node {$*$}  (psiA) ;

	\draw (psiAB) edge [myedge] node [myedgenode,above] {$a$}  (psiT);
	\draw (psiANB) edge [myedge] node [myedgenode,above] {$\neg a$}  (psiT);
\end{tikzpicture}
\caption{Graphical representation of the contraction operator \( \contraction \) for \( \epistemicSpace \) given in Example \ref{ex:es_contraction_operator}. Nodes are the epistemic states of \( \epistemicSpace \). Edges represent the behaviour of \( \contraction \); there is an edge from \( \Psi_1 \) to \( \Psi_2 \) with label \( x \) if \( x\equiv\alpha \) implies \( \Psi_2 = \Psi_1 \contraction \alpha \) hold. We use \( *  \) as placeholder label that stand for all \( x \) that are not explicitly mentioned.}\label{fig:ex:es_contraction_operator}
\end{figure}
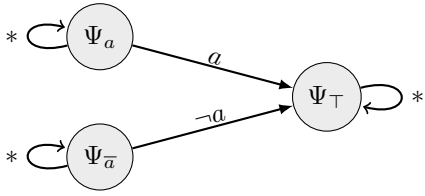
Note that presupposing existence of \( \contraction \) in the beginning of Definition~\ref{def:contraction_fullmeet_maxichoice}, guarantees implicitly the existence of the epistemic states used by \( \contraction \), as otherwise, \( \contraction \) would not exist.
Hence, in Definition~\ref{def:contraction_fullmeet_maxichoice}, we do not have to deal with the question of realizability. 

For a maxichoice contraction operators for epistemic states, for each epistemic state \( \Psi \), the linear order \( \ll_{\Psi} \) could be different. We denote maxichoice contraction operators that are based, globally, on a single linear order as \emph{linear contraction operators}:
\begin{definition}\label{def:con_linear}
    Let \( \epistemicSpace = \langle \setAllES,\beliefSymb \rangle  \) be an epistemic space.
    A \( \contraction \) contraction operator for \( \epistemicSpace \) is called a \emph{linear contraction operator for \( \epistemicSpace \)} if \( \contraction \) is a maxichoice contraction and there is a linear order \( \ll \) on \( \Omega \) such that for all \( \Psi\in\setAllES \) holds \( {\ll_{\Psi}} =  {\ll} \) for \( \ll_{\Psi} \) from Definition~\ref{def:contraction_fullmeet_maxichoice}.
\end{definition}
Next, we consider an example.
\begin{example}\label{ex:es_contraction_operator} 
    Let \( \Sigma=\{ a \} \) and thus \( \Omega=\{a,\overline{a}\} \).
    Consider the following epistemic space \( \epistemicSpace^{\mathrm{E}\ref{ex:es_contraction_operator}} = \langle \setAllES,\beliefSymb \rangle  \) with \( \setAllES=\{ \Psi_{a}, \Psi_{\overline{a}},  \Psi_{\top} \} \) and \( \beliefSymb \) is given by:
    \begin{align*}
        \modelsOf{\Psi_{a}} & = \{a\} & \modelsOf{\Psi_{\overline{a}}} & = \{\overline{a}\} & \modelsOf{\Psi_{\top}} & = \Omega &
    \end{align*}
    The set \( \setAllES \) contains only epistemic states for which the beliefs are consistent, i.e., \( \modelsOf{\Psi}\neq\emptyset \) for all \( \Psi\in\setAllES \).
    Let \( \contraction \) be a belief change operator for \( \epistemicSpace^{\mathrm{E}\ref{ex:es_contraction_operator}}  \) given as follows for all \( \Psi\in\setAllES \) and \( \alpha\in\logicLanguage \):
    \begin{equation*}
        \Psi\contraction\alpha = \begin{cases}
            \Psi_{\top} & \ksIF \alpha\in\beliefsOf{\Psi} \\
            \Psi & \ksOtherwise
        \end{cases}
    \end{equation*}
    	For a graphical representation of \( \contraction \) consider Figure \ref{fig:ex:es_contraction_operator}.
    	One can check that \( \contraction \) is a linear contraction operator, because the linear order \( \ll \) given by \( a \ll \overline{a} \) gives rise to this operator.
\end{example}

The following proposition points out that, when AGM contraction is realizable, i.e., there is an AGM contraction operators, then there is also an instance of all the concrete contraction operators considered here.
\begin{proposition}\label{col:es_contraction_linear}
	Let \( \epistemicSpace \) be an epistemic space. The following statements are equivalent:
	\begin{enumerate}[(I)]
		\item AGM contraction is realizable in \( \epistemicSpace \).
        \item There exists a linear contraction operator \( \contraction \) for \( \epistemicSpace \).
        \item\hspace{-0.5em}There exists a maxichoice contraction operator \( \contraction \) for~\( \epistemicSpace \).
        \item There exists a full meet contraction operator \( \contraction \) for \( \epistemicSpace \).
	\end{enumerate}
\end{proposition}
In the next section, we will see that for revision the situation is much more complex, as realizablity does not carry over between different types of revision operators.
 
\section{Realizability of Revision Operators}
\label{sec:realization_revision}
Now, we will determine the conditions of realizability of revision operators and specific types of revision operators.
Realizability of AGM revision is defined analogously to Definition~\ref{def:realizContraction}.
\begin{definition}[Revision Realizability]
	We say that \emph{AGM revision is realizable} in an epistemic space \( \epistemicSpace \), if there exists an AGM contraction operator for \( \epistemicSpace \).
\end{definition}
The next theorem characterizes exactly those epistemic spaces that permit existence of revision operators.
\begin{theorem}\label{prop:min_es_for_agmrevision}
	Let \( \epistemicSpace = \langle \setAllES , \beliefSymb \rangle  \) be an epistemic space. AGM revision is realizable in \( \epistemicSpace \) if and only if  \( \epistemicSpace \) satisfies:
	\begin{description}
		\item[\normalfont(\textlabel{ZR1}{pstl:AGM-ES1})]\hspace{-0.6em} For all \( \omega \,{\in}\, \Omega \) there is some \( \Psi_{\!\omega} \,{\in}\, \setAllES \) with \( {\modelsOf{\Psi_{\!\omega}} \,{=}\, \{\omega\} }\).
		\item[\normalfont(\textlabel{ZR2}{pstl:AGM-ES2})]\hspace{-0.6em} For all \( \Psi\in\setAllES \) and for all \( M\subseteq \modelsOf{\Psi} \) there exists \( \Psi_{M}\in\setAllES \) with \( \modelsOf{\Psi_{M}}=M \).
	\end{description}
\end{theorem}
The postulates \eqref{pstl:AGM-ES1} and \eqref{pstl:AGM-ES2} characterize epistemic spaces which have revision operators.
Figuratively, \eqref{pstl:AGM-ES1} describes  that the agent is capable of imagining that the \enquote{real world} is as described by exactly one interpretation; 
\eqref{pstl:AGM-ES2} describes that if an agent can consider a set of interpretations as possible, then this is true for every subset thereof.
A consequence of Theorem~\ref{prop:min_es_for_agmrevision} is that there are epistemic spaces where no revision operator exist at all.

As for contraction, we define types of revision operators, for which we will also consider the realizability.
\begin{definition}\label{def:rev_fullmeet_maxichoice}
    Let \( \epistemicSpace = \langle \setAllES,\beliefSymb \rangle  \) be an epistemic space and let \( \revision \) be a revision operator for \( \epistemicSpace \).
    	We say \( \revision \) is a \emph{full meet revision operator for \( \epistemicSpace \)} if for each \( \Psi\in\setAllES \) and \( \alpha\in\logicLanguage \) holds:
            \begin{equation*} %
            \modelsOf{\Psi\revision\alpha}=\begin{cases}
                \modelsOf{\Psi}\cap\modelsOf{\alpha}  & \ksIF \modelsOf{\Psi}\cap\modelsOf{\alpha} \neq \emptyset\\
                \modelsOf{\alpha}  & \ksOtherwise
            \end{cases}
    \end{equation*}
    	We say  \( \revision \) is a \emph{maxichoice revision operator for \( \epistemicSpace \)} if for each \( \Psi \,{\in}\, \setAllES \) exists a linear order \( \ll_{\!\Psi} \) such that for all \( \alpha \,{\in}\, \logicLanguage \) holds: 
        \begin{equation*} %
            \modelsOf{\Psi\revision\alpha} = \begin{cases}
                \modelsOf{\Psi}\cap\modelsOf{\alpha}  & \ksIF \modelsOf{\Psi}\cap\modelsOf{\alpha} \neq \emptyset\\
                \minOf{\modelsOf{{\alpha}}}{\ll_\Psi} & \ksOtherwise
            \end{cases}
        \end{equation*}
\end{definition}
We denote maxichoice revision operators that are based, globally, on a single linear order as \emph{linear revision operators}.
\begin{definition}\label{def:rev_linear}
	Let \( \epistemicSpace = \langle \setAllES,\beliefSymb \rangle  \) be an epistemic space.
	A \( \revision \) revision operator for \( \epistemicSpace \) is called a \emph{linear revision operator for \( \epistemicSpace \)} if \( \revision \) is a maxichoice revision and there is a linear order \( \ll \) on \( \Omega \) such that for all \( \Psi\in\setAllES \) holds \( {\ll_{\Psi}} =  {\ll} \) for \( \ll_{\Psi} \) from Definition~\ref{def:rev_fullmeet_maxichoice}.
\end{definition}
We consider an example epistemic space with a maxichoice revision operator, yet no full meet revision operators.
\begin{example}\label{ex:es_revision_operator}
	Let \( \Sigma=\{ a,b \} \) and thus \( \Omega=\{ {a}{b}, \overline{a}{b}, {a}\overline{b}, \overline{a}\overline{b} \} \).
	Consider the following epistemic space \( \epistemicSpace^{\mathrm{E}\ref{ex:es_revision_operator}} = \langle \setAllES,\beliefSymb \rangle  \) with \( \setAllES=\{ \Psi_{\omega} \mid \omega\in\Omega \}\cup\{ \Psi_\bot \} \) and \( \beliefSymb \) is given by:
	\begin{align*}
		\modelsOf{\Psi_{\omega}} & = \{\omega\} & \modelsOf{\Psi_{\bot}} & = \emptyset
	\end{align*}
	Let \( \ll_1 \) and \( \ll_2 \) be two linear orders on \( \Omega \) given by:
	\begin{align*}
		&  {a}{b}  \ll_1 {a}\overline{b} \ll_1 \overline{a}{b} \ll_1 \overline{a}\overline{b} &
		&  {a}{b}  \ll_2 \overline{a}{b} \ll_2 {a}\overline{b} \ll_2 \overline{a}\overline{b} 
	\end{align*}
	We employ \( \ll_1 \) and \( \ll_2 \) to define a maxichoice revision operator \( \revision \) for \( \epistemicSpace \) given as follows for all \( \omega\in\Omega \) and \( \alpha\in\logicLanguage \):
	\begin{align*}
		\Psi_{\omega}\revision\alpha & = \begin{cases}
			\Psi_{\bot} & \ksIF \modelsOf{\alpha}=\emptyset \\
			\Psi_{\omega} & \ksIF \omega\in\modelsOf{\alpha} \\
			\Psi_{\minOf{\modelsOf{\alpha}}{\ll_1}} & \ksOtherwise
		\end{cases} \\
		\Psi_{\bot}\revision\alpha & = \begin{cases}
			\Psi_{\bot} &  \ksIF \modelsOf{\alpha}=\emptyset \\
			\Psi_{\minOf{\modelsOf{\alpha}}{\ll_2}} & \ksOtherwise
		\end{cases}
	\end{align*}
	    For a graphical representation of \( \revision \) consider Figure \ref{fig:ex:es_revision_operator}.
To see that there is no full meet revision operator for \( \epistemicSpace^{\mathrm{E}\ref{ex:es_revision_operator}} \), we consider the epistemic state \( \Psi_{\bot} \) and the formula \( \alpha = b \). If \( \revision_{\mathrm{fm}} \) were a full meet revision operator for \( \epistemicSpace^{\mathrm{E}\ref{ex:es_revision_operator}} \), then we would obtain \( \modelsOf{\Psi_{\bot} \revision_{\mathrm{fm}} \alpha} = \{ ab, \overline{a}{b} \} \). However, this is impossible as there is no epistemic state \( \Psi \in \setAllES \) with \( \modelsOf{\Psi}= \{ ab, \overline{a}{b} \} \).
\end{example}
Example~\ref{ex:es_revision_operator} is a witness for the following observation.
\begin{proposition}\label{prop:nofullmeet}
	AGM revision is realizable in \( \epistemicSpace^{\mathrm{E}\ref{ex:es_revision_operator}} \), yet there exist no full meet revision for \( \epistemicSpace^{\mathrm{E}\ref{ex:es_revision_operator}} \).
\end{proposition}

The next proposition exactly determines the realizability of linear revision operators, maxichoice revision operators, and of full meet revision operators.
\begin{proposition}\label{prop:realizefullmeet_linear}
Let \( \epistemicSpace \) be an epistemic space. The following statements hold:
    \begin{enumerate}[\normalfont(I)]
        \item AGM revision is realizable in \( \epistemicSpace \) if and only if there exists a linear revision operator \( \revision \) for \( \epistemicSpace \).
        \item  AGM revision is realizable in \( \epistemicSpace \) if and only if there exists a maxichoice revision operator \( \revision \) for \( \epistemicSpace \).
        \item  There exists a full meet revision operator \( \revision \) for \( \epistemicSpace \) if and only if the following postulate is satisfied by \( \epistemicSpace \):\\[-0.5em]
        \begin{description}
            \item[\hspace{-1.6em}\normalfont(\textlabel{Unbiased}{pstl:unbiased})]\hspace{-0.6em} For all belief sets \( B \subseteq \logicLanguage \) there exists an epistemic state \( \Psi_{\!B} \in \setAllES \) with \( {\beliefsOf{\Psi_{\!B}} = B }\).
        \end{description}
    \end{enumerate}    
\end{proposition}
Due to Proposition~\ref{prop:realizefullmeet_linear} (I), linear revision operators are a canonical type of revision operators. 
Clearly, because linear revision operators are maxichoice revision operators, we obtain Proposition~\ref{prop:realizefullmeet_linear} (II) from Proposition~\ref{prop:realizefullmeet_linear} (I).
Moreover, Proposition~\ref{prop:realizefullmeet_linear} (III) shows that full meet revision operators exist only when the agent is able to believe in any belief sets.
\begin{figure}[t]
\begin{center}		
    \resizebox{0.9\columnwidth}{!}{\begin{tikzpicture}
            \usetikzlibrary{decorations.pathreplacing}
            \tikzstyle{esnode}=[circle,draw,fill=gray!15,text=black,minimum width=2.5em,minimum height=1.75em]
            \tikzstyle{myedge}=[->,thick,-{latex}]
            
            \tikzstyle{myedgenode}=[inner sep=0.2ex,sloped]

            \node (psiPAPB) [esnode,anchor=west] at (0,0) {\(\Psi_{{a}{b}}\)};
            \node (psiNAPB) [esnode,anchor=west] at (3,0) {\(\Psi_{\overline{a}{b}}\)};
            \node (psiPANB) [esnode,anchor=west] at (0,3) {\(\Psi_{{a}\overline{b}}\)};
            \node (psiNANB) [esnode,anchor=west] at (3,3) {\(\Psi_{\overline{a}\overline{b}}\)};
            \node (psiBOT) [esnode,anchor=west] at (6,1.5) {\(\Psi_{\top}\)};
            
            \draw (psiPAPB) edge [loop left,thick] node {$ $}  (psiPABP) ;
            \draw (psiPANB) edge [loop left,thick] node {$ $}  (psiPANB) ;
            \draw (psiNAPB) edge [loop right,thick] node {$ $}  (psiPANB) ;
            \draw (psiNANB) edge [loop right,thick] node {$ $}  (psiPANB) ;
            \draw (psiBOT) edge [loop right,thick] node {$\bot$}  (psiBOT) ;

            \draw (psiPAPB) edge [myedge,bend left=15] node [myedgenode,above] {$ $}  (psiPANB);
            \draw (psiPAPB) edge [myedge,bend left=15] node [myedgenode,above] {$ $}  (psiNAPB);
            \draw (psiPAPB) edge [myedge,bend left=15] node [myedgenode,above] {$ $}  (psiNANB);
            
            \draw (psiPANB) edge [myedge,bend left=15] node [myedgenode,below] {$ $}  (psiPAPB);
            \draw (psiPANB) edge [myedge,bend left=15] node [myedgenode,above] {$ $}  (psiNAPB);
            \draw (psiPANB) edge [myedge,bend left=15] node [myedgenode,above] {$ $}  (psiNANB);
            
            \draw (psiNAPB) edge [myedge,bend left=15] node [myedgenode,below] {$ $}  (psiPAPB);
            \draw (psiNAPB) edge [myedge,bend left=15] node [myedgenode,below] {$ $}  (psiPANB);
            \draw (psiNAPB) edge [myedge,bend left=15] node [myedgenode,above] {$ $}  (psiNANB);
            
            \draw (psiNANB) edge [myedge,bend left=15] node [myedgenode,below] {$ $}  (psiPAPB);
            \draw (psiNANB) edge [myedge,bend left=15] node [myedgenode,below] {$ $}  (psiPANB);
            \draw (psiNANB) edge [myedge,bend left=15] node [myedgenode,below] {$ $}  (psiNAPB);
            
            \draw [decorate,thick,decoration={brace,amplitude=10pt,mirror,raise=4pt},yshift=0pt]
            (4.75,-0.55) -- (4.75,3.5) node (mynode) [black,midway,xshift=12.5pt,inner sep=0] {};
            \draw (mynode) edge [myedge] node [myedgenode,above] {$\bot$}  (psiBOT);
            
            \draw [thick]  (psiBOT) --    ([yshift=0.6cm]psiBOT |- psiPANB.north) -- ([yshift=0.6cm]psiPANB.north) edge [myedge] (psiPANB) ;
            \draw [thick]  (psiBOT) --    ([yshift=0.3cm,xshift=-0.3cm]psiBOT |- psiNANB.north) -- ([yshift=0.3cm]psiNANB.north) edge [myedge] (psiNANB) ;
            
            \draw [thick]  (psiBOT) --    ([yshift=-0.6cm]psiBOT |- psiPAPB.south) -- ([yshift=-0.6cm]psiPAPB.south) edge [myedge] (psiPAPB) ;
            \draw [thick]  (psiBOT) --  node [above,sloped] {\(   \)}  ([yshift=-0.3cm,xshift=-0.3cm]psiBOT |- psiNAPB.south) -- ([yshift=-0.3cm]psiNAPB.south) edge [myedge] (psiNAPB) ;
    \end{tikzpicture}}
\end{center}
    \caption{Graphical representation of the revision operator \( \revision \) for \( \epistemicSpace \) given in Example \ref{ex:es_revision_operator}. Due to space reasons we omitted most labels.}\label{fig:ex:es_revision_operator}
 \end{figure}
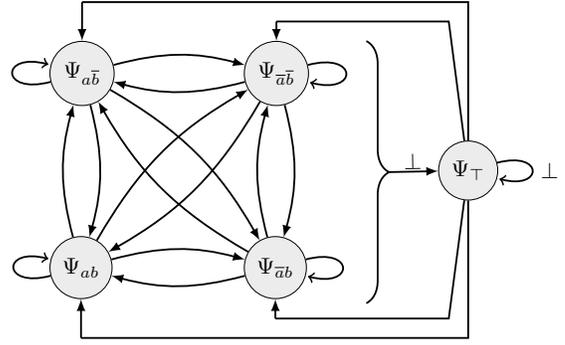 
\section{Consequences for the Interdefinability of Revision and Contraction}
\label{sec:interaction}
In the classical AGM belief change framework, revision and contraction are interdefinable \cite{KS_FermeHansson2018}, via the Levi-identity \cite{KS_Levi1977} and Harper-identity \cite{KS_Harper1976}.
As observed at first by \citeauthor{KS_KoniecznyPinoPerez2017} (\citeyear{KS_KoniecznyPinoPerez2017}), revision and contraction operators over epistemic spaces are not interdefinable in general.
\begin{proposition}[{\citeauthor{KS_KoniecznyPinoPerez2017}}, {\citeyear{KS_KoniecznyPinoPerez2017}}]\label{prop:noleviharper}
	There is an epistemic space \( \epistemicSpace \) such that there exist more\footnote{\label{ftn:moreoperator}By cardinality; the set of all revision operators for \( \epistemicSpace \) has strictly more elements than the set of all contraction operators for \( \epistemicSpace \).} revision operators for \( \epistemicSpace \) than contraction operators for~\( \epistemicSpace \).
\end{proposition}

The results of this paper allow further remarks to the matter of interdefinability of revision and contraction.
First, observe that \( \epistemicSpace^{\mathrm{E}\ref{ex:es_revision_operator}} \) from Example~\ref{ex:es_revision_operator} contains the epistemic state \( \Psi_{ab} \) with \( \modelsOf{\Psi_{ab}} = \{ ab \} \), yet there is \emph{no} epistemic state \( \Psi\in\setAllES \) such that \( \modelsOf{\Psi}=\{ ab, {a}\overline{b} \} \) holds. Hence, due to Theorem~\ref{prop:min_es_for_agmcontraction}, there is no contraction operator for \( \epistemicSpace^{\mathrm{E}\ref{ex:es_revision_operator}} \); showing that the situation can be more drastic than Proposition~\ref{prop:noleviharper} suggests.
\begin{proposition}\label{prop:noleviharper_realiz}
     AGM contraction is not realizable in \( \epistemicSpace^{\mathrm{E}\ref{ex:es_revision_operator}} \), but 
    AGM revision is realizable in \( \epistemicSpace^{\mathrm{E}\ref{ex:es_revision_operator}} \).
\end{proposition}
Moreover, in specific epistemic spaces a situation that is dual to Proposition~\ref{prop:noleviharper} appears, as witnessed by Example~\ref{ex:es_contraction_operator}.
\begin{proposition}\label{prop:noleviharper_morecontraction}
	AGM revision is not realizable in \( \epistemicSpace^{\mathrm{E}\ref{ex:es_contraction_operator}} \), but 
	AGM contraction is realizable in \( \epistemicSpace^{\mathrm{E}\ref{ex:es_contraction_operator}} \).
\end{proposition}
We obtain the following dual to Proposition~\ref{prop:noleviharper}.
\begin{corollary}
	There is an epistemic space \( \epistemicSpace \) such that there exist more contraction operators for \( \epistemicSpace \) than revision operators for \( \epistemicSpace \).
\end{corollary}

\section{Conclusion and Discussion}
In this paper, we considered the problem of realizability for AGM revision and AGM contraction in epistemic spaces. 
Next, we highlight and discuss some of this paper's results, outline a plan for future work, and identify open problems.

\myparagraph{Realizability and Canonical Operators}
The main results, Theorem~\ref{prop:min_es_for_agmcontraction} and Theorem~\ref{prop:min_es_for_agmrevision}, show that AGM revision and AGM contraction, are only realizable in specific epistemic spaces.
Our results support that for epistemic spaces, linear contraction and linear revision are conical instances, as they are guaranteed to exist when the respective kind of belief change is realizable at all (Proposition~\ref{col:es_contraction_linear} and Proposition~\ref{prop:realizefullmeet_linear}).
We want to highlight that the observation that linear revisions are a canonical form of revision for epistemic spaces is converse to the observations made when generalizing AGM revision to arbitrary Tarskian logics, where full meet revision operators always exist and, thus, are canonical instances for AGM revision \cite{KS_FalakhRudolphSauerwald2022}.
Not that the setting considered here is more general than in \cite{KS_FalakhRudolphSauerwald2022}, as, e.g., a closure under conjunction over the available epistemic states is not assumed.
Our results on realizability are also different from those of \citeauthor{KS_RibeiroWassermannFlourisAntoniou2013} (\citeyear{KS_RibeiroWassermannFlourisAntoniou2013}), who investigated which logics AGM revision and AGM contraction operators exist. Still, they did not consider the restrictions given by considering epistemic spaces, which leads to different results.

\myparagraph{Potential Application}
We predict that our realizability results can be used, for instance, to checking beforehand, whether AGM revision or AGM contraction can be applied out of the box in a specific domain. Another application are algorithms that compute, for a given epistemic space \( \epistemicSpace \), whether a revision or contraction operator for \( \epistemicSpace \) exists.

\myparagraph{Linear Contraction and Linear Revision}
Note that linear contraction and linear revision are novel types of operators, which make sense when considering the setting of epistemic spaces. 
Before us, \citeauthor{HerzigKoniecznyPerrussel2003} (\citeyear{HerzigKoniecznyPerrussel2003}) already employed linear orders to define belief change operators. However, the difference is that they did not employ linear orders to specify the global behaviour of operators as we did here in Section~\ref{sec:realization_contraction} and Section~\ref{sec:realization_revision}; their definition corresponds to what is here called maxichoice.

\myparagraph{Future Work and Open Problems}
In future work, we extend our investigations of revision and contraction constrained by iteration postulates \cite{KS_SauerwaldKern-IsbernerBeierle2020,KS_SchwindKoniecznyPinoPerez2022,KS_DarwichePearl1997}.
We are also planning to consider the exact relation of belief change in epistemic spaces to hyper-intensional belief revision \cite{KS_SouzaWassermann2022,KS_Berto2019,KS_OezguenBerto2021} and to belief change in subclassical logics, for which abstract logic \cite{KS_LewitzkaBrunner2009} will be useful, as well as the work by Flouris \cite{KS_Flouris2006,KS_FlourisPlexousakis2006} and results in the understanding of revision in Horn logic \cite{KS_DelgrandePeppas2015}.
As highlighted in this paper, another interesting problem to study is the interrelation of revision and contraction in epistemic spaces. The interested reader may consult \citeauthor{KS_KoniecznyPinoPerez2017} (\citeyear{KS_KoniecznyPinoPerez2017}) and \citeauthor{KS_BoothChandler2019}~{(\citeyear{KS_BoothChandler2019})} as a starting point.
An open problem for the area of (iterated) belief change is to consider more expressive logics and explore what are reasonable belief changes in these settings.
The notion of epistemic spaces might be useful for such endeavours. We expect that our realizability results carry over to more expressive settings, and we plan to check whether this conjecture holds.
\pagebreak[3]
\appendix
\section*{Acknowledgments}
We would like to thank the reviewers for their detailed comments and valuable feedback. 
Especially, we like to thank the reviews for drawing connections to other areas, e.g., for drawing the connection to abstract logics.
This research is supported by the German Research Association (DFG), project number 465447331.
\pagebreak[3]

\bibliographystyle{kr}
\bibliography{bibliography}

\clearpage
\section{Supplementary Material}

\newenvironment{repeatprop}[1]{\smallskip\par\noindent\textbf{Proposition~\ref{#1}}.\itshape}{\par}
\newenvironment{repeatthm}[1]{\smallskip\par\noindent\textbf{Theorem~\ref{#1}}.\itshape}{\par}
\newenvironment{repeatcorollary}[1]{\smallskip\par\noindent\textbf{Corollary~\ref{#1}}.\itshape}{\par}

\subsection{Background}
In this section, we present some background results that we need for the proofs in the next section.

A semantic characterization of AGM contraction is epistemic spaces is by faithful assignments.
\begin{definition}[Adapted, {\citeauthor{KS_DarwichePearl1997}}, {\citeyear{KS_DarwichePearl1997}}]
	Let \( \epistemicSpace = \langle \setAllES,\beliefSymb \rangle  \) be an epistemic space.
	A faithful assignment for \( \epistemicSpace \) is a function \( \Psi \mapsto {\leq_{\Psi}} \) 
	that maps each epistemic state \( \Psi\in\setAllES \) to a total preorder \( {\leq_{\Psi}} \subseteq \Omega\times\Omega \) over  \( \Omega \) 
	such that \( \modelsOf{\Psi}\neq\emptyset \) implies \( \modelsOf{\Psi}=\minOf{\Omega}{\leq_{\Psi}} \).
\end{definition}
We link contraction operators and faithful assignments by a notion of compatibility.
\begin{definition}[Adapted, \citeauthor{KS_FalakhRudolphSauerwald2022}, {\citeyear{KS_FalakhRudolphSauerwald2022}}]
    Let \( \epistemicSpace = \langle \setAllES,\beliefSymb \rangle  \) be an epistemic space and let \( \contraction \) be a belief change operator  for \( \epistemicSpace \).
    A faithful assignment \( \Psi \mapsto {\leq_{\Psi}} \) (for \( \epistemicSpace \)) is called \emph{contraction-compatible} with \( \contraction \) if the following holds:
    \begin{equation}\tag{contraction-compatible}\label{eq:repr_es_contraction}
        \modelsOf{\Psi \contraction \alpha} = \modelsOf{\Psi} \cup \min(\modelsOf{\negOf{\alpha}},\leq_{\Psi})
    \end{equation}
\end{definition}

A characterization in terms of total preorders on epistemic states is given by the following proposition.
\begin{proposition}[Adapted, {\citeauthor{KS_KoniecznyPinoPerez2017}}, {\citeyear{KS_KoniecznyPinoPerez2017}}]\label{prop:es_contraction}
    Let \( \epistemicSpace = \langle \setAllES,\beliefSymb \rangle  \) be an epistemic space and let \( \contraction \) be a belief change operator  for \( \epistemicSpace \).
    We have that \( \contraction \) is a contraction operator for \( \epistemicSpace \) if and only if there is a faithful assignment $ \Psi\mapsto {\leq_\Psi} $ for \( \epistemicSpace \) which is \ref{eq:repr_es_contraction} with \( \contraction \).
\end{proposition}

\subsection{Proofs}
In this section, we present proofs for the novel results in our paper.

\begin{repeatthm}{prop:min_es_for_agmcontraction}
	Let \( \epistemicSpace = \langle \setAllES,\beliefSymb \rangle  \) be an epistemic space.
	AGM contraction is realizable in \( \epistemicSpace \) if and only if \( \epistemicSpace \) satisfies:
	\begin{description}
		\item[\normalfont(\ref{pstl:AGM-ES3})] For each \( \Psi\in\setAllES \) and for each \( M \subseteq \Omega \) with \( \modelsOf{\Psi}\subseteq M \)\\\null\hspace{2.25em} there exists \( \Psi_{M}\in\setAllES \) such that \( \modelsOf{\Psi_{M}}=M \).
	\end{description}
\end{repeatthm}
\begin{proof} 
    We start by showing that the existence of \( \contraction \) implies satisfaction of \eqref{pstl:AGM-ES3}.
    For each \( \omega\in\Omega \) let  \( \varphi_{\overline{\omega}} \) be a formula such that \( \modelsOf{\varphi_{\overline{\omega}}}=\Omega\setminus\{\omega\} \) holds. 
    Let \( \Psi\in\setAllES \) be an arbitrary epistemic state and \( M \subseteq \Omega \) be such that \( \modelsOf{\Psi}\subsetneq M \) holds.
    In the following let \( M\setminus\modelsOf{\Psi} =\{ \omega_1,\ldots,\omega_n \} \).
    Using Proposition~\ref{prop:es_contraction} we obtain that \( \Psi_M=\left(((\Psi\contraction\varphi_{\overline{\omega_1}})\contraction\varphi_{\overline{\omega_2}} ) \ldots\right) \contraction \varphi_{\overline{\omega_n}} \) is an epistemic state with \( \modelsOf{\Psi_M}=M \).
    
    For the remaining direction, let \( \setAllES \) be a (non-empty) set of epistemic states that satisfies \eqref{pstl:AGM-ES3}.
    We construct a linear contraction operator for \( \setAllES \).
    Let \( \ll \) be an arbitrary linear order on \( \Omega \), and for each \( \omega \in \Omega \) let \( \Psi_{\{\omega\}} \) denote a uniquely chosen epistemic state from \( \setAllES \) such that \( \modelsOf{\Psi_{\{\omega\}}}=\modelsOf{\Psi}\cup\{\omega\} \). 
    Let \( \contraction \) be the belief change operator  for \( \setAllES \) given as follows:
    \begin{equation*}
        \Psi \contraction \alpha = \begin{cases}
            \Psi &\ksIF \alpha\notin\beliefsOf{\Psi} \\
            \Psi_{\minOf{\modelsOf{\negOf{\alpha}}}{\ll}} & \ksOtherwise
        \end{cases}
    \end{equation*}
    Because \( \setAllES \) satisfies \eqref{pstl:AGM-ES3}, we have that for each \( \Psi\in\setAllES \) and \( \alpha\in\logicLanguage \) the epistemic state \( \Psi\contraction\alpha \) exists, i.e., \( \Psi\contraction\alpha\in\setAllES \).
    Now let \( \Psi\mapsto{\leq_{\Psi}} \) be defined for each \( \Psi\in\setAllES \) as follows:
    \begin{equation*}
        {\leq_{\Psi}} = \left({\ll}\setminus (\Omega\times\modelsOf{\Psi})\right)\,\cup\, (\modelsOf{\Psi}\times\Omega).
    \end{equation*}
    Indeed, \( \Psi\mapsto{\leq_{\Psi}} \) is a faithful assignment. Moreover, \( \Psi\mapsto{\leq_{\Psi}} \) is \ref{eq:repr_es_contraction} with \( \contraction \).
    By Proposition~\ref{prop:es_contraction}, we obtain that \( \contraction \) is a contraction operator for \( \setAllES \).
\end{proof}

\begin{repeatprop}{col:es_contraction_linear}
	Let \( \epistemicSpace \) be an epistemic space. The following statements are equivalent:
	\begin{enumerate}[(I)]
		\item AGM contraction is realizable in \( \epistemicSpace \).
		\item There exists a linear contraction operator \( \contraction \) for \( \epistemicSpace \).
		\item\hspace{-0.5em}There exists a maxichoice contraction operator \( \contraction \) for~\( \epistemicSpace \).
		\item There exists a full meet contraction operator \( \contraction \) for \( \epistemicSpace \).
	\end{enumerate}
\end{repeatprop}
\begin{proof}
   Clearly, if one of the Statements~(II)--(IV) holds, we immediately obtain Statement~(I).
   For Statement (I) implies Statement~(II), observe that the operator \( \contraction \) constructed in the proof of Theorem~\ref{prop:min_es_for_agmcontraction} is a linear contraction operator.
   Moreover, because any linear contraction operator is also a maxichoice contraction, we obtain that Statement~(II) implies Statement~(III).
   We show that Statement~(I) implies Statement~(IV).
   Let \( \epistemicSpace=\tuple{\setAllES,\beliefSymb} \) be an epistemic space that satisfies \eqref{pstl:AGM-ES3}.
   We construct a full meet contraction operator for \( \epistemicSpace \).
   For each \( M \subseteq \Omega \) let \( \Psi_{M} \) denote a uniquely chosen epistemic state from \( \setAllES \) such that \( \modelsOf{\Psi_{M}}=\modelsOf{\Psi}\cup M \), which exists due to the courtesy of \eqref{pstl:AGM-ES3}. Let \( \contraction \) be the belief change operator  for \( \setAllES \) given as follows:
   \begin{equation*}
       \Psi \contraction \alpha = \begin{cases}
           \Psi &\ksIF \alpha\notin\beliefsOf{\Psi} \\
           \Psi_{\modelsOf{\negOf{\alpha}}} & \ksOtherwise
       \end{cases}
   \end{equation*}
   The operator \( \contraction \) is AGM contraction operator for \( \epistemicSpace \) because when using the flat order \( {\leq_{\Psi}} = \Omega\times\Omega \) for every epistemic state \( \Psi\in\setAllES \), one obtains that \( \Psi\mapsto {\leq_{\Psi}} \) is a faithful assignment \ref{eq:repr_es_contraction} with \( \contraction \).
   
   Given the above-mentioned insights on \( \contraction \), one can easily check that the conditions for full meet are given in Definition~\ref{def:contraction_fullmeet_maxichoice} are met by \( \contraction \).
\end{proof}

\begin{repeatthm}{prop:min_es_for_agmrevision}
	Let \( \epistemicSpace = \langle \setAllES , \beliefSymb \rangle  \) be an epistemic space. AGM revision is realizable in \( \epistemicSpace \) if and only if  \( \epistemicSpace \) satisfies:
	\begin{description}
		\item[\normalfont(\ref{pstl:AGM-ES1})]\hspace{-0.6em} For all \( \omega \,{\in}\, \Omega \) there is some \( \Psi_{\!\omega} \,{\in}\, \setAllES \) with \( {\modelsOf{\Psi_{\!\omega}} \,{=}\, \{\omega\} }\).
		\item[\normalfont(\ref{pstl:AGM-ES2})]\hspace{-0.6em} For all \( \Psi\in\setAllES \) and for all \( M\subseteq \modelsOf{\Psi} \) there exists \( \Psi_{M}\in\setAllES \) with \( \modelsOf{\Psi_{M}}=M \).
	\end{description}
\end{repeatthm}
\begin{proof} We start by showing that the existence of \( \revision \) implies satisfaction of \eqref{pstl:AGM-ES1} and \eqref{pstl:AGM-ES2}.
    From satisfaction of \eqref{pstl:R1} and \eqref{pstl:R3} we obtain satisfaction of \eqref{pstl:AGM-ES1}. Therefore, pick an arbitrary epistemic state \( \Psi\in\setAllES \) and a formula \( \varphi_\omega \) such that \( \modelsOf{\varphi_\omega}=\{\omega\} \). 
    Because of \eqref{pstl:R1} and \eqref{pstl:R3} there exists some epistemic state \( \Psi_\omega\in\setAllES \) such that \( \modelsOf{\Psi_\omega}=\{\omega\} \) and \( \Psi\revision\varphi_\omega=\Psi_\omega \).
    For satisfaction of \eqref{pstl:AGM-ES2} consider some \( \Psi\in\setAllES \). 
    For each \( M\subseteq\modelsOf{\Psi} \) let \( \varphi_M \) be such that \( \modelsOf{\varphi_M}=M \). Clearly, because of \eqref{pstl:R2}, revision of \( \Psi \) by \( \varphi_M \) yields an epistemic state \( \Psi\revision\varphi_M \) such that \( \modelsOf{\Psi\revision\varphi_M}=M \) is satisfied.
    
    For the remaining direction, let \( \setAllES \) be a (non-empty) set of epistemic states that satisfies \eqref{pstl:AGM-ES1} and \eqref{pstl:AGM-ES2}.
    Let \( \ll \) be an arbitrary linear order on \( \Omega \) and for each \( M \subseteq \Omega \) let \( \Psi_M \) denote a uniquely chosen epistemic state from \( \setAllES \) such that \( \modelsOf{\Psi_M}=M \) (as far as such an epistemic state \( \Psi_M \) exists in \( \setAllES \)). 
    We construct a revision operator \( \revision \) as follows:
    \begin{equation*}
        \Psi\revision \alpha = \begin{cases}
            \Psi_{\modelsOf{\Psi}\cap \modelsOf{\alpha}} &\ksIF \modelsOf{\Psi}\cap \modelsOf{\alpha}\neq\emptyset \\
            \Psi_{\minOf{\modelsOf{\alpha}}{\ll}} & \ksOtherwise
        \end{cases}
    \end{equation*}
    Observe that, because of \eqref{pstl:AGM-ES2}, the epistemic state \( \Psi_{\modelsOf{\Psi}\cap \modelsOf{\alpha}} \) exists whenever \( \modelsOf{\Psi}\cap \modelsOf{\alpha}\neq\emptyset \) holds.
    Moreover, because \( \ll \) is a linear order, we have, for every consistent \( \alpha \), that \( \minOf{\modelsOf{\alpha}}{\ll} \) is a singleton set, and thus, by \eqref{pstl:AGM-ES1} the epistemic state \( \Psi_{\minOf{\modelsOf{\alpha}}{\ll}} \) exists. For inconsistent \( \alpha \), the epistemic state \( \Psi_{\minOf{\modelsOf{\alpha}}{\ll}} \)  exists by \eqref{pstl:AGM-ES2} (and non-emptiness of \( \setAllES \)). In summary, for every \( \Psi\in\setAllES \) and \( \alpha\in\logicLanguage \) the epistemic state \( \Psi\revision\alpha \) exists.
    
    We show satisfaction of \eqref{pstl:R1}--\eqref{pstl:R6} by \( \revision \).
    The postulates \eqref{pstl:R1}--\eqref{pstl:R3} are immediately satisfied due to the construction of \( \revision \).
    For \eqref{pstl:R4} observe that the behaviour of \( \revision \) by input \( \alpha \) is only defined via models of \( \Psi \) and \( \alpha \).
    Consequently, equivalent input formulas are treated equivalently.
    If \( \modelsOf{\Psi\revision\alpha}\cap\modelsOf{\beta} \) is inconsistent, then \eqref{pstl:R5} and \eqref{pstl:R6} are immediately satisfied.
    In the following assume that \( \modelsOf{\Psi\revision\alpha}\cap\modelsOf{\beta} \) is consistent.
    For the case of \( \modelsOf{\Psi\revision\alpha} \cap \modelsOf{\Psi}\neq\emptyset \), we obtain \eqref{pstl:R5} and \eqref{pstl:R6} from \eqref{pstl:R2}.
    For the case of \( \modelsOf{\Psi\revision\alpha} \cap \modelsOf{\Psi}=\emptyset \), we have by construction that \( \modelsOf{\Psi\revision\alpha} \) is a singleton set, i.e., \( \modelsOf{\Psi\revision\alpha}=\{\omega\} \).
    Consequently, from consistence of \( \modelsOf{\Psi\revision\alpha}\cap\modelsOf{\beta} \) we obtain \( \omega\in \modelsOf{\beta} \).
    From minimality of \( \omega \) with respect to \( \ll \), we obtain that \( \minOf{\modelsOf{\alpha}\cap\modelsOf{\beta}}{\ll}=\{\omega\} \). 
    We obtain satisfaction of \eqref{pstl:R5} and \eqref{pstl:R6}.
\end{proof}

\begin{repeatprop}{prop:nofullmeet}
    AGM revision is realizable in \( \epistemicSpace^{\mathrm{E}\ref{ex:es_revision_operator}} \), yet there exist no full meet revision for \( \epistemicSpace^{\mathrm{E}\ref{ex:es_revision_operator}} \).
\end{repeatprop}
\begin{proof}
    AGM revision is realizable in \( \epistemicSpace^{\mathrm{E}\ref{ex:es_revision_operator}} \) is witnessed by \( \revision \) in Example~\ref{ex:es_revision_operator}. To show the second part of the statement, we make use of Proposition~\ref{prop:realizefullmeet_linear}, and show that \eqref{pstl:unbiased} is violated by \( \epistemicSpace^{\mathrm{E}\ref{ex:es_revision_operator}} \). Clearly, this is the case, because there is no epistemic state \( \Psi \) in \( \epistemicSpace^{\mathrm{E}\ref{ex:es_revision_operator}} \) such that \( \modelsOf{\Psi}=\Omega \) holds, i.e., that \( \beliefsOf{\Psi} = \Cn(\{a\lor \negOf{a}\}) \) holds.
\end{proof}

\begin{repeatprop}{prop:realizefullmeet_linear}
	Let \( \epistemicSpace \) be an epistemic space. The following statements hold:
    \begin{enumerate}[\normalfont(I)]
        \item AGM revision is realizable in \( \epistemicSpace \) if and only if there exists a linear revision operator \( \revision \) for \( \epistemicSpace \).
        \item  AGM revision is realizable in \( \epistemicSpace \) if and only if there exists a maxichoice revision operator \( \revision \) for \( \epistemicSpace \).
        \item  There exists a full meet revision operator \( \revision \) for \( \epistemicSpace \) if and only if the following postulate is satisfied by \( \epistemicSpace \):\\[-0.5em]
        \begin{description}
            \item[\hspace{-1.6em}\normalfont(\ref{pstl:unbiased})]\hspace{-0.6em} For all belief sets \( B \subseteq \logicLanguage \) there exists an epistemic state \( \Psi_{\!B} \in \setAllES \) with \( {\beliefsOf{\Psi_{\!B}} = B }\).
        \end{description}
    \end{enumerate}       
\end{repeatprop}
\begin{proof}    
    We consider the Statements (I)--(III) independently.
    \begin{itemize}
        \item[]\hspace{-1em}\emph{\normalfont(I).} We start with Statement (I).
        If there exists a linear revision operator \( \revision \) for \( \epistemicSpace \), then there exists a revision operator \( \revision \) for \( \epistemicSpace \).
        For the other direction, consider the proof of Theorem \ref{prop:min_es_for_agmrevision}: the construction for \( \revision \) in this proof is a linear revision operator.
        
        \item[]\hspace{-1em}\emph{\normalfont(II).} For Statement (II), recall that every linear revision operator is also a maxichoice revision operator.
        Hence, from Statement (I), we obtain one direction of Statement (II).
        If there exists a maxichoice revision operator \( \revision \) for \( \epistemicSpace \), then there exists a revision operator \( \revision \) for \( \epistemicSpace \).
        
        \item[]\hspace{-1em}\emph{\normalfont(III).} We continue with Statement (III).
        If there exists a full meet revision operator \( \revision \) for \( \epistemicSpace \), then by Theorem~\ref{prop:min_es_for_agmrevision}, \eqref{pstl:AGM-ES1} and \eqref{pstl:AGM-ES2} are satisfied by \( \epistemicSpace \).
        Now observe that, as \( \revision \) exists, for each \( \Psi\in\setAllES \) and each \( M \) with \( M\subseteq \Omega\setminus\modelsOf{\Psi}   \) there exists some epistemic state \( \Psi_M \) with \( \modelsOf{\Psi_M}=M \). This is because \( \revision \) is a full meet revision.
        From \eqref{pstl:AGM-ES2} we obtain some epistemic state with \( \Psi_\bot\in\setAllES \) with \( \modelsOf{\Psi_\bot}=\emptyset \). 
        Consequently, for each \( M\subseteq \Omega\setminus\modelsOf{\Psi_\bot}=\Omega \) there exists some epistemic state \( \Psi_M \) with \( \modelsOf{\Psi_M}=M \).
        Now recall that \( \beliefsOf{\Psi} \)  is fully determined by \( \modelsOf{\Psi_M} \), i.e., \( \beliefsOf{\Psi_M}=\{ \alpha\in\logicLanguage \mid M \models \alpha \} \).
        We obtain that \eqref{pstl:unbiased} is satisfied by \( \epistemicSpace \) as for \( M\subseteq \Omega \) there is one epistemic state  \( \Psi_M \) such that \( \modelsOf{\Psi_{M}}=M \).
        
        For the other direction, assume that \eqref{pstl:unbiased} is satisfied by \( \epistemicSpace \).
        Let \( \Psi_M \) denote be a unique epistemic state from \( \setAllES \) such that \( \modelsOf{\Psi_M}=M \) for every \( M \subseteq \Omega \).
        We define the belief change operator \( \revision \) by:
        \begin{equation*}
            \Psi \revision \alpha = \begin{cases}
                \Psi_{\modelsOf{\Psi}\cap\modelsOf{\alpha}} &  \ksIF \modelsOf{\Psi}\cap\modelsOf{\alpha}\neq\emptyset\\
                \Psi_{\modelsOf{\alpha}} & \ksOtherwise
            \end{cases}
        \end{equation*}
        One can verify that \( \revision \) is a full meet revision operator.\qedhere
    \end{itemize}
\end{proof}

\begin{repeatprop}{prop:noleviharper_realiz}
	AGM contraction is not realizable in \( \epistemicSpace^{\mathrm{E}\ref{ex:es_revision_operator}} \), but 
	AGM revision is realizable in \( \epistemicSpace^{\mathrm{E}\ref{ex:es_revision_operator}} \).
\end{repeatprop}
\begin{proof}
    Consider \( \epistemicSpace^{\mathrm{E}\ref{ex:es_revision_operator}} \) from Example~\ref{ex:es_revision_operator}, by Theorem~\ref{prop:min_es_for_agmrevision}, there exists a revision operator for \( \epistemicSpace^{\mathrm{E}\ref{ex:es_revision_operator}} \).
    By Theorem~\ref{prop:min_es_for_agmcontraction} there exists no contraction operator for \( \epistemicSpace^{\mathrm{E}\ref{ex:es_revision_operator}} \). 
\end{proof}

\begin{repeatprop}{prop:noleviharper_morecontraction}
	AGM revision is not realizable in \( \epistemicSpace^{\mathrm{E}\ref{ex:es_contraction_operator}} \), but 
	AGM contraction is realizable in \( \epistemicSpace^{\mathrm{E}\ref{ex:es_contraction_operator}} \).
\end{repeatprop}
\begin{proof}
    Consider \( \epistemicSpace^{\mathrm{E}\ref{ex:es_contraction_operator}} \) and \( \contraction \) from Example \ref{ex:es_contraction_operator}.
    Recall that \( \contraction \) is a contraction operator for \( \setAllES \).
    By Theorem~\ref{prop:min_es_for_agmrevision} there exists no revision operator for \( \setAllES \). 
\end{proof}

\end{document}